\documentclass{article}

\usepackage[final,nonatbib]{./neurips_robustbayes21}

\usepackage[utf8]{inputenc} % allow utf-8 input
\usepackage[T1]{fontenc}    % use 8-bit T1 fonts
\usepackage[style=numeric,
            backend=biber,
            giveninits=true,
            uniquename=init,
            url=false,
            doi=false,
            eprint=false,
            isbn=false,
            maxbibnames=10,
            sorting=nyt,
            natbib=true,
            sortcites=no,
            hyperref=true]{biblatex}

\usepackage{hyperref}      % hyperlinks
\hypersetup{
  colorlinks=true,
  linkcolor=blue,
  filecolor=magenta,
  urlcolor=cyan,
  citecolor=blue
}
\usepackage{url}            % simple URL typesetting
\usepackage{booktabs}       % professional-quality tables
\usepackage{nicefrac}       % compact symbols for 1/2, etc.

\usepackage{wrapfig}
\usepackage[pdftex]{graphicx}
\usepackage{breqn}
\usepackage{subfiles}
\usepackage{enumitem}
\usepackage{placeins}

\usepackage{microtype,marvosym}
\usepackage{amsmath,amssymb, amsfonts}

\usepackage{tikz,pgfplots}
\usepackage{booktabs}

\renewcommand{\d}{\:d}

\newcommand{\dd}{\mathrm{d}}

\renewcommand{\Re}{\mathbb{R}}

\newcommand{\one}{\mathbf{1}}

\newcommand{\N}{\mathcal{N}}

\newcommand{\Trans}{^{\intercal}}

\renewcommand{\det}{\operatorname{det}}

\newcommand{\q}{\quad}

\renewcommand{\vec}{\boldsymbol}

\renewcommand{\O}{\mathcal{O}}

\newcommand{\GP}{\mathcal{GP}}
\newcommand{\Id}{\vec{I}}

\newcommand{\bmu}{\boldsymbol{\mu}}
\newcommand{\bSigma}{\boldsymbol{\Sigma}}

\newcommand{\x}{\vec{x}}

\newcommand{\X}{\mathbb{X}}

\usepackage{colonequals}

\usepackage{nicefrac}
\usepackage{tensor}

\graphicspath{{./figs/}{figs/}}

\title{Invariant Priors for Bayesian Quadrature}

\author{%
  Masha Naslidnyk\thanks{Work done while at Amazon.com.} \\
  University College London\\
  \texttt{masha.naslidnyk.21@ucl.ac.uk} \\
  \And
  Javier Gonzalez\footnotemark[1]\\
  Microsoft Research Cambridge\\
  \texttt{gonzalez.javier@microsoft.com} \\
  \And
  Maren Mahsereci\footnotemark[1] \\
  University of T\"ubingen\\
  \texttt{maren.mahsereci@uni-tuebingen.de} \\
}

\begin{document}

\maketitle

\begin{abstract}
  Bayesian quadrature (BQ) is a model-based numerical integration method that is
  able to increase sample efficiency by encoding and leveraging known structure of the integration
  task at hand.
  In this paper, we explore priors that encode
  \emph{invariance} of the integrand under a set of
  bijective transformations in the input domain, in particular some
  unitary transformations, such as rotations, axis-flips, or point
  symmetries.
  We show initial results on superior performance in comparison to standard Bayesian quadrature on several
  synthetic and one real world application.
\end{abstract}

\section{Introduction}
\label{sec:introduction}

The numerical solution of an intractable integral
explicitly or implicitly influences a model's fidelity,
and the decision that is based upon it.
Examples are computing the expected outcome of a physical experiment,
and propagating the result in a pipeline;
or estimating the expected output of a computer simulation,
and then acting upon the solution;
but also, and somewhat more implicit,
computing the evidence of a probabilistic model, and relying on its predictive power.
In all those cases, integral solvers affect the outcome.

When solving an integral, we are often restricted in the number of
integrand evaluations (sample size $N$) which is framed as
an \emph{expensive} integration problem,
e.g., when evaluating the integrand equals a monetary or time investment.
In those cases, sample efficiency is key, and
Monte Carlo (MC) integration may yield estimators that
are high in variance \parencite{Rasmussen2003}.
Bayesian quadrature (BQ) \parencite{Larkin1972, Diaconis1988, OHagan1991} is a model-based numerical integration
method that is extremely suited for small samples sizes and has been shown to outperform MC methods on several,
especially low-dimensional tasks \parencite{Rasmussen2003}.
BQ places a surrogate model on the integrand $f$, usually a
Gaussian process (GP) \parencite{Rasmussen2006},
and then, since that is tractable, integrates the model in place of the real integrand.
Besides an improved estimator, BQ generally yields a univariate distribution over the integral value
that quantifies its epistemic uncertainty (for more details on the connection to probabilistic numerics,
please refer to \parencite{Hennig2015, Briol2019}).
The general reason for increased sample efficiency is that,
due to the model-based approach, known information
about the integrand can be encoded into the prior.

In this paper, we explore a set of priors for Bayesian quadrature that encode
invariance of the integrand $f$ under
input transformations, that is $f(\x)=f(T(\x))$ where
$T$ is a bijective map from and to the input domain of $f$.
In initial experiments, we explore a subset of
those transformations (axis flips), which increases the
sample efficiency in comparison to BQ with a non-invariant prior.

\paragraph{Related Work:}
The authors of \cite{Gunter2014} encode
non-negativity of $f$, generally true when integrating probability densities,
by modelling the square-root of the true integrand; \cite{ChaiGarnett19}
by modeling the logarithm.
Other works tailor the algorithm towards a specific application
\parencite{Osborne2012a, Osborne2012b, Ma2014, Xi2018, Gessner19}.
The authors of \cite{Karvonen2017a} exploit structure of the Gaussian process
regression equations and are able to reduce algorithmic complexity
(generally given by $\O(N^3)$ for GP inference) by
evaluating $f$ in specified locations.
\emph{Active} and \emph{adaptive} BQ \cite{Gessner19, Gunter2014, ChaiGarnett19, KanHen19}
propose a sequential acquisition strategy for $(\x_i, y_i)$ to obtain
most informative integrand evaluations.
Even without an active acquisition strategy, BQ may improve apon a
model-free approach as the BQ estimator explicitly dependents on the surrogate model \cite{Rasmussen2003}.

\section{Background}
\label{sec:background}

\paragraph{Setting}
We consider the problem of numerically approximating an intractable integral
$Z = \int_{ \Omega } f(\x) \pi(\x) \dd \x$, $f: \X \rightarrow \Re$,
where the integration domain $\Omega \subseteq \X\subseteq \Re^d $ may be either a Cartesian product of finite bounds $[l_1, u_1]\times...\times[l_d, u_d]$, or the entire $\Re^d$.
The integration measure $\pi(\x)$ is the Lebesgue measure $\pi(\x) \equiv 1$, or the normal density $\N(\x; \bmu, \bSigma)$ respectively.

\paragraph{Bayesian quadrature}

Bayesian quadrature (BQ) \parencite{Diaconis1988, OHagan1991} models the integrand $f$ with a Gaussian process $f(\x)\sim \GP(m(\x), k(\x, \x'))$ \parencite{Rasmussen2006} with mean function $m$ and kernel function $k$ that can be conditioned on integrand evaluations or 'data' $D = \{(\x_i, y_i)\ |\ i = 1..N\}$ \parencite{Hennig2015, Cockayne2017, Briol2019}.
Observations are exact, $y_i = f(\x_i)$, but if needed, Gaussian noise can be added.
The stacked observations are denoted as $Y\in\Re^{N}$, and the corresponding stacked locations as $X\in\Re^{N\times d}$. In a slight abuse of notation, we will refer to both the Gaussian process and the actual blackbox function as $f$.

With a distribution over $f$, we can obtain not only an estimator for $Z$, but a full posterior distribution which can subsequently be used in decision making or uncertainty analysis \parencite{Hennig2015}. Due to the closeness property of Gaussians under affine transformations, the integral $Z\sim \N(\mu_Z, \sigma^2_Z)$ is 1D-normally distributed with scalar mean $\mu_Z$ and variance $\sigma^2_Z$ (please refer to Appendix~A for full formulas).
In essence, BQ requires the computation of the kernel mean $qK^X := \int_{\Omega}k(\x, X)\pi(\x)\dd \x$ evaluated at $X$, the prior variance $qKq :=\iint_{\Omega\times\Omega}k(\x, \x')\pi(\x)\pi(\x')\dd\x\dd\x'$, and the integral over the prior mean $\int_{\Omega}m(\x)\pi(\x)\dd\x$.
These can be computed analytically for certain kernels, such as the well-known RBF-kernel $k(\x, \x') = \theta^2\exp(-\nicefrac{\|\x-\x'\|^2}{(2\lambda^2)})$, and a zero mean function which we will use throughout the paper.

\paragraph{Invariant Gaussian processes}
\label{sec:invar-gauss-proc}
A function $f:\X \rightarrow \Re$
is invariant under a bijective transformation $T:\X\rightarrow \X$ if $f(T(\x)) = f(\x)$ holds for all $\x$ in $\X$.
Consider now a function which is invariant under $J$ transformations $T_i$, $i=1, ..., J$.
The set of those transformations is called $G_f:=\{T | f(\x) = f(T(\x)) \text{ for all } \x \in\X\}$  and forms a group.
The set $\mathcal{A}(\x):=\{T(\x) | \text{ for all } T\in G_f\}$ is defined as the set of invariant locations induced by $\x$.
Thus we can introduce a function $g: \X \rightarrow \Re$ such that
$f(\x) = \sum_{\tilde{\x}\in \mathcal{A}(\x)}g(\tilde{\x})$ \parencite{NIPS2018_8199}.
We can see that any point $\tilde{\x}\in A(\x)$ induces the identical set $A(\tilde{\x}) = A(\x)$, thus $f(\tilde{\x}) = f(\x)$ for all $\tilde{\x}\in A(\x)$
(note that $G_f$ includes the identity transform, the trivial invariance for all functions).
We are free to model $g$ as a Gaussian process, such that $g\sim\GP(m_g, k_g)$ with mean function $m_g$ and kernel function $k_g$.
Thus, $f$ is also a Gaussian process $f\sim\GP(m_f, k_f)$, as $f$ is a linear combination of Gaussian distributed values. The mean function $m_f$ and kernel function $k_f$ can be easily derived \parencite{NIPS2018_8199}:
\begin{equation}
  \label{eq:5}
  \begin{split}
    m_f(\x) =\sum_{\tilde{\x}\in \mathcal{A}(\x)}m_g(\tilde{\x}),\q
    k_f(\x, \x') = \sum_{\substack{\tilde{\x}\in \mathcal{A}(\x) \\ \tilde{\x}' \in \mathcal{A}(\x')}}  k_g(\tilde{\x}, \tilde{\x}').
  \end{split}
\end{equation}
GP regression on $f$ is straightforward,
as Eq.~\ref{eq:5} simply defines another positive definite
kernel.

\section{BQ priors for invariant integrands}
\label{sec:invar-bayes-quadr}

We would now like to use the invariant Gaussian process as prior model for BQ.
It is less straightforward, however, to analytically extract the
push-forward measure on the integral value $Z$.
This is because $\x$ over which the integral is performed, is now transformed (possibly nonlinearly) by $T$.
More precisely, the integrals for $i, j=1,\dots, J$ required are
\begin{equation*}
  qK_{ij}^X : = \int_{\Omega} k(T_i(\x), T_j(X))\pi(\x)\dd\x,~\text{and}~
  qKq_{ij}: =
\iint_{\Omega\times \Omega} k_{g}(T_i(\x), T_j(\x')) \pi(\x)\pi(\x')\dd\x \dd\x'.
\end{equation*}
Even if the original kernel integrals (without the transformations $T_i$) are known analytically, these ones
might be arbitrarily complicated or even intractable.
However, the distribution over the integral value $Z\sim
\N(\mu_Z, \sigma^2_Z)$ remains Gaussian, as we are still
integrating the Gaussian process on $f$
(we are not warping the
distribution; we are merely distorting the input space).
Please refer to Appendix~C for full expressions of the posterior integral mean and variance.

\begin{wrapfigure}{r}{0.4\textwidth}
  \centering
  \includegraphics[width=0.4\textwidth]{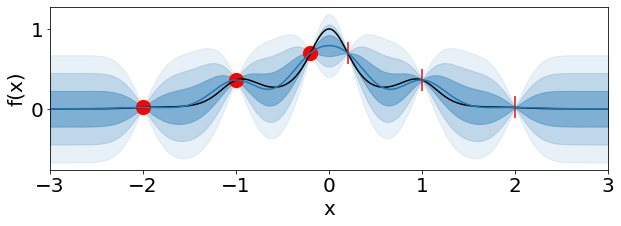}
  \caption{
    Invariant GP. Details in text.
  }
  \label{fig:1d_symmetric_illustration}
\end{wrapfigure}
An illustration of a simple invariance in 1D is shown in
Figure~\ref{fig:1d_symmetric_illustration}. The true integrand in black is axis-symmetric which
can be represented by the transformation $T_1(\x) = -\x$. The set $G_f$
thus contains the identity transform $T_0$ as well as $T_1$. In the
figure, the three observed points are marked as red dots.
The invariant locations implied by $T_1$ are marked as red vertical bars.
Even though these points are not observed, the model's uncertainty shrinks at and around the implied locations.
We want to highlight that using a standard GP with additional `fake' observations at the invariant locations
would cubicly increase the algorithmic cost of the method according to $\O(J^3N^3)$.
The invariant GP has complexity $\O(N^3)$.
Furthermore, samples from the invariant GP obey the invariant property as well,
but samples from a standard GP with `fake' observations do not.
In the next section, we
consider a subset of orthogonal transformations which
allow for analytic computation of the required kernel integrals.

\subsection{Orthogonal transformations}
\label{sec:linear}
Consider that $G_f$ contains orthogonal transformations only, that is
$T_i(\x) = Q_i\x$, $i=1,\dots J$ with $Q_i$ a square matrix, and $Q_i\Trans Q_i =
\Id$. Orthogonal transforms, among others, include rotations, point-
and axis-projections.
For the RBF-kernel introduced in Section~\ref{sec:introduction},  we can write the
kernel mean for the $i$th and $j$th transform as:
\begin{equation}
  \label{eq:1}
  qK_{ij}(\x) = qK(Q_i\Trans Q_j \x)
\end{equation}
where $qK$ is the kernel mean of the RBF-kernel without the
transforms, evaluated  at $\vec{s}_{ij}=Q_i\Trans Q_j \x$. The equality is
easily shown by re-arranging terms and using the orthogonal property
of the $Q_i$.
Eq.~\ref{eq:1} might in itself be valuable as it allows for the
computation of the mean estimator of the integral value
$\mu_Z^X$. Unfortunately though, it is less straightforward to compute the
double integrals $qKq_{ij}$ required for the integral variance $\sigma^{2, X}_Z$, and this may only be
analytically possible for a subset of orthogonal transforms.
One of those subsets are axis- and point symmetries, discussed in
the next section.

\subsection{Axis-flips and point-symmetries}
\label{sec:axis-align-symm}
For orthogonal transformations described in Section~\ref{sec:linear}, and
with a change of variables $Q_i\Trans \circ Q_j:\Omega \rightarrow
\X$, $Q_i\Trans Q_j \x \mapsto \vec{s}_{ij}$, the
variances $qKq_{ij}$ can be written as
\begin{equation}
  \label{eq:3}
  qKq_{ij} = |\det^{-1}(Q_i\Trans Q_j)|\int_{Q_i\Trans \circ
    Q_j(\Omega)}qK(\vec{s}_{ij})\pi(\vec{s}_{ij})\dd \vec{s}_{ij}.
\end{equation}
Even if the original integral $\int_{\Omega}
qK(\x)\pi(\x)\dd \x$ is analytic as is the case for the RBF-kernel,
the linear transform $Q_i\Trans Q_j$ generally distorts the integration
domain as well as $\pi(\x)$,
yielding a possibly intractable integral.
One way to circumvent this is to impose that the transform $Q_i\Trans
Q_j$ needs to be diagonal for all $i, j=1,...J$. There
might be other scenarios where the integral remains analytic, here we only study this obvious case.
This characteristic is fulfilled by $Q_i$ that encode
axis- and point-symmetries, i.e, when the function $f$ is invariant
under flipping the sign of one or more axis.
Hence, the
transformations are of the form $T_i(\x) = Q_i\x$ with
$Q_{i}^{kl}=c_i^k\delta_{kl}$ and $c_i^k\in\{-1, 1\}$. One
or multiple of sign-flips can be present separately in the set $G_f$.
An illustration of several sets $\mathcal{A}(\x)$ for different $G_f$, as well as
the derivation of the necessary integrals
are provided in Appendix~B.

We will call a method encoding invariaces `invariant-BQ' in contrast to `standard-BQ'.
We highlight that the integration problem itself needs not be invariant (only $f$ does), as
$\pi(\x)$ or $\Omega$ need not be invariant.
Further, the implied invariant observations influence the posterior consistently across symmetric planes
or points as can be seen from Figure~\ref{fig:1d_symmetric_illustration}.

\section{Experiments}
\label{sec:symmetric-fx}

We provide initial experiments that indicate increased sample efficiency of invariant-BQ in comparison to
standard-BQ, on a set of synthetic examples, and a Fourier optics problem.

\subsection{Synthetic examples}
\label{sec:upperc-exampl}
\begin{wrapfigure}{r}{0.5\textwidth}
  \centering
  \includegraphics[width=0.5\textwidth]{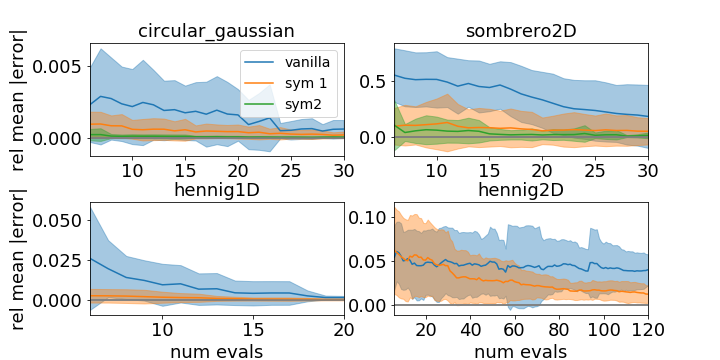}
  \caption{Average performance of standard-BQ and invariant-BQ on 4 synthetic examples.
  }
  \label{fig:symm_bq_performance_avergaes}
\end{wrapfigure}
We consider a collection of four simple functions implemented as quadrature
test function in the \textit{Emukit}
Python library \parencite{emukit2019}, all of which exhibit point- and/or axis symmetries
(Appendix~B.1 for definitions and illustrations).
We apply invariant-BQ and integrate with respect to the Lebesgue
measure $\pi(x) \equiv 1$ on a finite domain $\Omega = [-3, 3]^d$ for
each function, and compare the performance to standard-BQ.

Figure~\ref{fig:symm_bq_performance_avergaes} shows the relative mean absolute
error between the mean estimator $\mu_Z$ and the
true value of the integral, across 10 runs with different random seeds
(Appendix~D for experimental details).
Not only does invariant-BQ (orange for point-, green for additional axis-symmetry) outperform the
standard model (blue)
on average at each step, it is also more stable as the estimate varies
less between runs.
Performance for a single run showing $\mu_Z$ and $\sigma_Z$ is plotted in Figure~\ref{fig:symm_bq_performance_single_run} in
Appendix~D.
Figures~\ref{fig:symm_bq_performance_avergaes_app} and
\ref{fig:symm_bq_performance_single_run_app}
show the same experiment with optimal kernel
hyperparameters $\lambda$ and $\theta$; the performance gain is even more
apparent there.
As a second experiment, we intgerate the same functions w.r.t. an isotropic Gaussian measure with density $\pi(\x)=\N(\x; \vec{b},  l^2\Id)$, mean $\vec{b}=\one$ and variance $l^2=1$.
As $\pi(\x)$ is not invariant, neither is the product $f(\x)\pi(\x)$.
The results are similar to the ones dicussed, and shown in Figure~\ref{fig:symm_bq_performance_avergaes_gauss} in Appendix~D.

\subsection{Point spread function}
\label{sec:upperc-spre-funct}
\begin{wrapfigure}{r}{0.5\textwidth}
  \centering
    \includegraphics[width=0.5\textwidth]{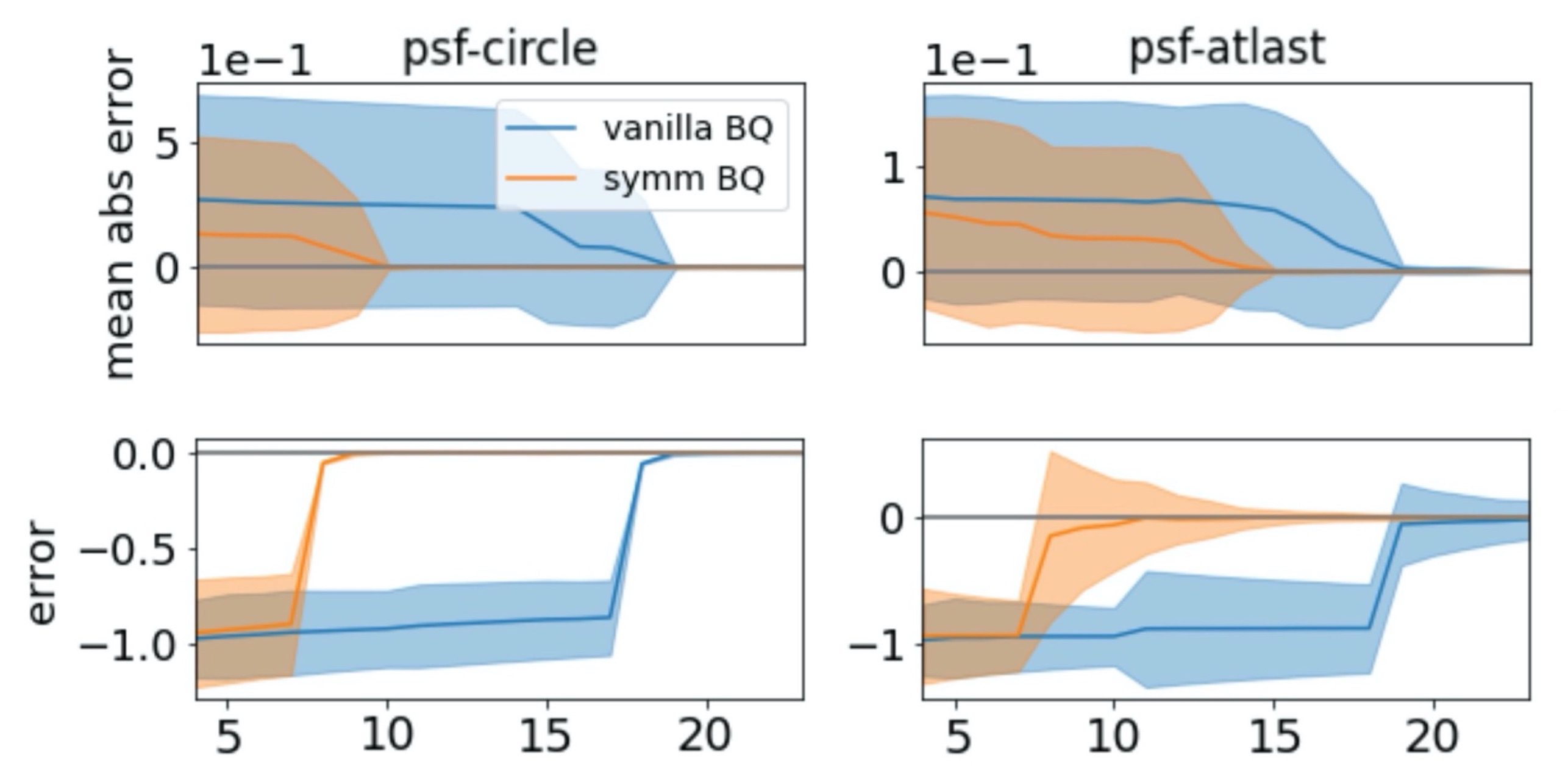}\\
    \caption{
      Average performance (t) vs a single run (b) on two PSFs;
      circular (l), ATLAST (r).
      }
  \label{fig:symm_bq_performance_psf}
\end{wrapfigure}
We apply invariant-BQ to
approximate the integral of two point spread functions (PSFs). In Fourier
optics, the point spread function is used to model the distribution of
light in the image of a point formed by an optical system
\parencite{artal2017handbook, mahajan1998optical}.
See Appendix~E for a description of the ATLAST and circular pupil used here.
We use the \textit{poppy}
Python library \parencite{poppy} to generate pupil functions and their respective PSFs.
The integral of the PSF is interpreted as the total power of the light
source in the image, and used to normalise the PSF so it may be
interpreted as a probability distribution.
We compute the PSF for both the circular and the ATLAST pupil, both point-symmetric.
The results are shown in Figure~\ref{fig:symm_bq_performance_psf}.
Invariant-BQ consistently outperforms standard-BQ here.

\section{Discussion \& Conclusion}
\label{conclusion}

We proposed `invariant-BQ', that leverages a GP prior encoding invariance of the integrand under a fixed set of input transformations.
The integration measure and domain need not be invariant.
We presented initial promising experiments that show improved sample efficiency over standard-BQ for a subset of transformations that allow for analytic GP inference.
In future work it would be desirable to test invariant-BQ on a more complete set of experimental settings and baselines, and to study the effect on decision making and predictive model fidelity.

\clearpage

\printbibliography[title={\normalsize References}]

\clearpage
\appendix

\begin{center}
{\Large ---Supplementary Material---}

{\large Invariant Priors for Bayesian Quadrature}
\end{center}

\vspace{1cm}

\section*{A: Formulas for Bayesian quadrature}

Recall that we consider the problem of numerically approximating the integral
$Z = \int_{ \Omega } f(\x) \pi(\x) \d \x$, $ f: \X \rightarrow \Re$,
with integration domain $\Omega \subseteq \X\subseteq \Re^d $ either a Cartesian product of
finite bounds $[l_1, u_1]\times...\times[l_d, u_d]$, or the entire $\Re^d$.
The integration measure $\pi(\x)$ is the Lebesgue measure $\pi(\x) \equiv 1$, or the normal density $\N(\x; \bmu, \bSigma)$ respectively.

Recall that the integrand $f$ is modelled with a Gaussian process $f(\x)\sim \GP(m(\x), k(\x, \x'))$ with mean function $m$ and kernel function $k$ that can be conditioned on integrand evaluations or 'data' $D = \{(\x_i, y_i)\ |\ i = 1..N\}$.
The stacked observations are denoted as $Y\in\Re^{N}$, and the corresponding stacked locations as $X\in\Re^{N\times d}$.
Then, the integral $Z\sim \N(\mu_Z, \sigma^2_Z)$ is 1D-normally distributed with scalar mean $\mu_Z$ and variance $\sigma^2_Z$:
\begin{equation}
\label{int_mean_var}
  \begin{split}
\mu_Z &= \int_{ \Omega }  m(\x) \pi(\x) \d \x + qK^XG^{-1}\Delta\\
\sigma^2_Z &= qKq - qK^XG^{-1}\left[qK^{X}\right]^{\intercal},
  \end{split}
\end{equation}
where $G\in\Re^{N\times N}$ is the kernel Gram matrix $G_{nm} = k(\x_n, \x_m)$, and $\Delta=Y-m(X)\in\Re^N$ is the residual.
Eq.~\ref{int_mean_var} requires the computation of the kernel mean $qK^X := \int_{\Omega}k(\x, X)\pi(\x)\dd \x\in\Re^{1\times N}$ evaluated at $X$, the scalar prior variance $qKq :=\iint_{\Omega\times\Omega}k(\x, \x')\pi(\x)\pi(\x')\dd\x\dd\x'$, as well as the integral over the prior mean $\int_{\Omega}m(\x)\pi(\x)\dd\x$.
The kernel integrals cannot always be computed analytically. They can for certain kernels, such as the well-known squared exponential/ RBF-kernel $k(\x, \x') = \theta^2\exp(-\nicefrac{\|\x-\x'\|^2}{(2\lambda^2)})$, which is parametrized by its scalar variance $\theta^2$ and lengthscale $\lambda$. The integral over the prior mean function can be solved e.g., for a constant mean $m(\x) = M$.
These are the prior kernel we are considering in the paper.

\FloatBarrier

\section*{B: Illustration of invariances in 2D}
\begin{figure}
  \centering
  \includegraphics[scale=0.1]{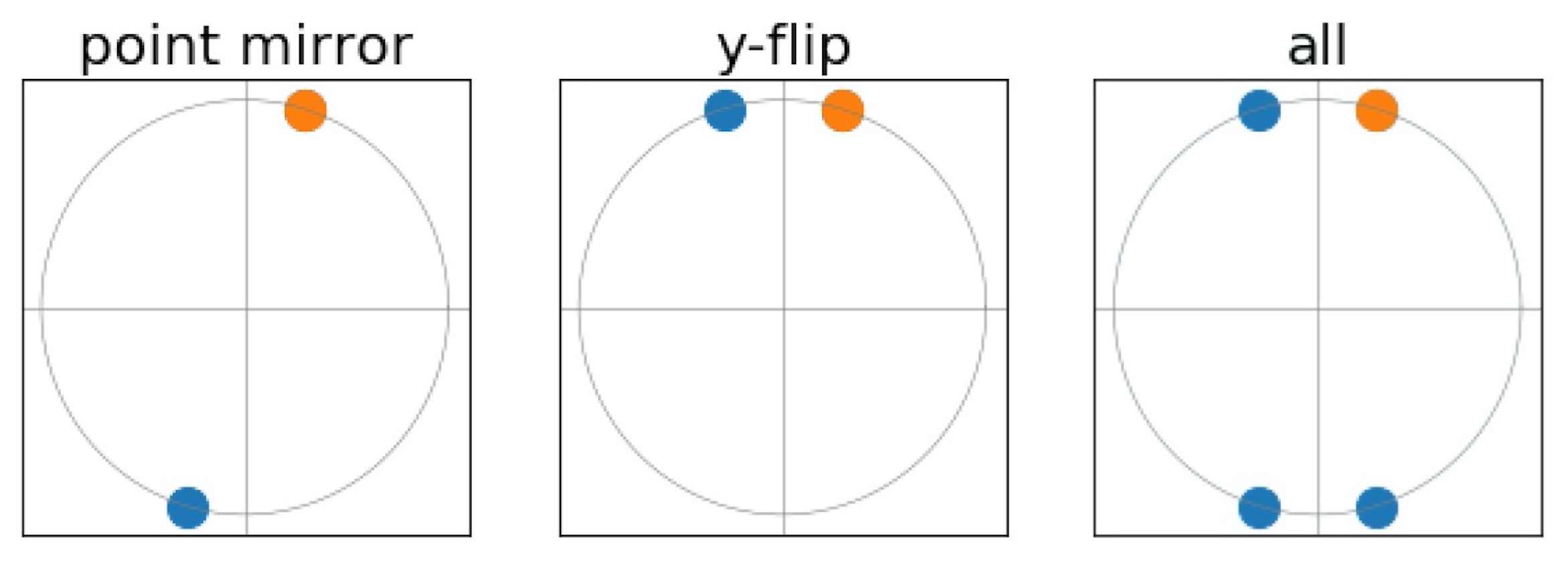}
  \caption{Set $\mathcal{A}(\x)$ for point symmetry
    w.r.t. origin (left), y-axis flip (middle), and their combination (right).
    The point $\x=[0.3, 1]\Trans$ (orange) is same for all sets; the induced points
    are shown in blue. The size of set $G_f$ is $J=2, 2$, and $4$ respectively.}
  \label{fig:Q_func}
\end{figure}
Consider axis- and point-symmetries as in Section~\ref{sec:axis-align-symm}.
The integration domain is transformed according to $Q_i\Trans \circ Q_j (\Omega)$, which
equals to flipping the integration bounds of dimension $q$ according to
$(Q_i\Trans Q_j)_{qq} = c_i^q c_j^q \in\{-1, 1\}$; the flip happens when the value is
$-1$.
The determinant of $Q_i\Trans Q_j$ is either 1 or $-1$,
and thus its absolute value is $1$.
Figure~\ref{fig:1d_symmetric_illustration} illustrates the 1D case, where $G_f$ consist
only of $Q_0(x)=x$ and $Q_1(x)=-x$, thus $Q_0 Q_1= Q_1Q_0 = -1$, and
$Q_0Q_0 =Q_1Q_1 = 1$. In fact, for the special case of axis flips
considered in this section, it
is true that $Q_i\Trans Q_j = Q_j\Trans Q_j$ for all $i, j$.
We provide an illustation of several sets $\mathcal{A}(\x)$ for different $G_f$ for dimension $d=2$
in Figure~\ref{fig:Q_func}.

\FloatBarrier

\section*{C: Formulas for invariant BQ}

Consider an invariant GP as introduced in Section~\ref{sec:background}.
The distribution over the integral value $Z\sim
\N(\mu_Z, \sigma^2_Z)$ remains Gaussian for invariant BQ.
The posterior formulas for the mean and variance of the integral value are
\begin{equation}
  \label{eq:8}
  \begin{split}
    \mu_Z
    &= \sum_{i=1}^J\int_{\Omega} m_g^X(T_i(\x))\pi(\x) \dd\x\\
    \sigma^{2}_Z
    &= \sum_{i, j=1}^J \iint_{\Omega\times\Omega} k_g^X(T_i(\x), T_j(\x'))\pi(\x)\pi(\x') \dd\x \dd\x',
  \end{split}
\end{equation}
where the superscript $^X$ indicates that the Gaussian process $g$ is conditioned on the $N$ integrand observations $Y$ at locations $X$.
We expand the expressions above:
\begin{equation}
  \label{eq:10}
  \begin{split}
 \mu_Z
 &= \sum_{i=1}^J\int_{\Omega} m_{g}(T_i(\x))\pi(\x)\dd\x
 + \sum_{i, j=1}^JqK^X_{ij} G^{-1}\Delta\\
\sigma^{2}_Z
&= \sum_{i, j=1}^J qKq_{ij}
-  \sum_{i, j=1}^JqK^X_{ij} G^{-1}  \sum_{k, l=1}^JqK^{X\intercal}_{kl},
  \end{split}
\end{equation}
with $G_{nm}=\sum_{i,j=1}^Jk(T_i(\x_n), T_j(\x_m))$ the kernel Gram
matrix, and $qK_{ij}^X : = \int_{\Omega} k(T_i(\x), T_j(X))\pi(\x)\dd\x$, the kernel mean
corresponding to invariances $T_i$ and $T_j$ evaluated at $X$, and $qKq_{ij}: =
\iint_{\Omega\times \Omega} k_{g}(T_i(\x), T_j(\x')) \pi(\x)\pi(\x')\dd\x \dd\x'$ the prior variance
corresponding to invariances $T_i$ and $T_j$. Note that $qK_{ij}^X $ is
a row vector of size $N$, where $N$ is the number of
observations, $qKq_{ij}$ is a scalar.
Even if the (double) integrals of $qK(\x)$ and $k(\x, \x')$ are known analytically,
the integrals $qK_i^X$ and $qKq_{ij}$ might be arbitrarily complicated or even intractable.

\section*{C: The synthetic functions considered for invariant-BQ}
All functions are taken from the BQ test function folder of the Emukit library \parencite{emukit2019}. They are:

\begin{equation*}
  \begin{split}
 &\mathop{\text{hennig1D}}(x) = e^{-x^2 -\sin^2(3x)}\\
 &\mathop{\text{hennig2D}}(\x) = e^{-\sin(3\|\x\|^2) - \x^TS\x},\quad S = \begin{bmatrix} 1 & 0.5 \\ 0.5 & 1 \end{bmatrix}\\
  &\mathop{\text{circular\_gaussian}}(\x) = \frac{1}{2\pi \sigma^2} \|\x\|^2
  e^{-\frac{\left(\|\x\| - \mu\right)^2}{2 \sigma^2}} \\
  &\hspace{5cm}\ \mu\in\Re, \sigma >0\\
&\mathop{\text{sombrero2D}}(\x) =  \frac{\sin(\pi \|\x\| c)}{\pi \|\x\| c},\q c \in \Re_{+}.
\end{split}
\end{equation*}

\begin{figure}
  \centering
  \includegraphics[scale=0.25]{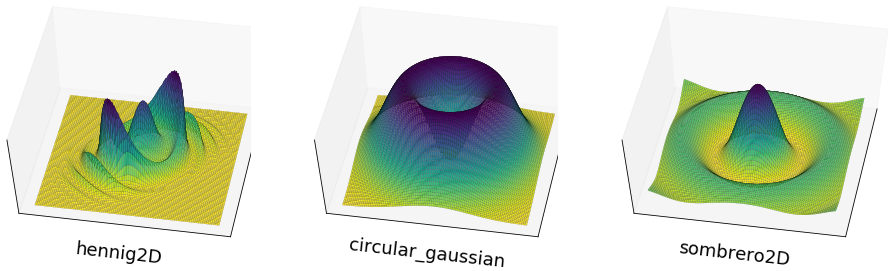}
  \caption{Three out of the four test functions, used in
    Figures~\ref{fig:symm_bq_performance_avergaes} and
    \ref{fig:symm_bq_performance_single_run}.
    Plot boundaries
    coincide with integration domain. All are point-symmetric
    w.r.t. the origin, 2 and 3 are also axis-symmetric.}
  \label{fig:symm_bq_test_functions}
\end{figure}

\FloatBarrier

\section*{D: Additional results of synthetic experiments in Section~\ref{sec:upperc-exampl}}

\begin{figure}[t]
  \centering
  \hspace*{-6mm}\includegraphics[scale=0.24]{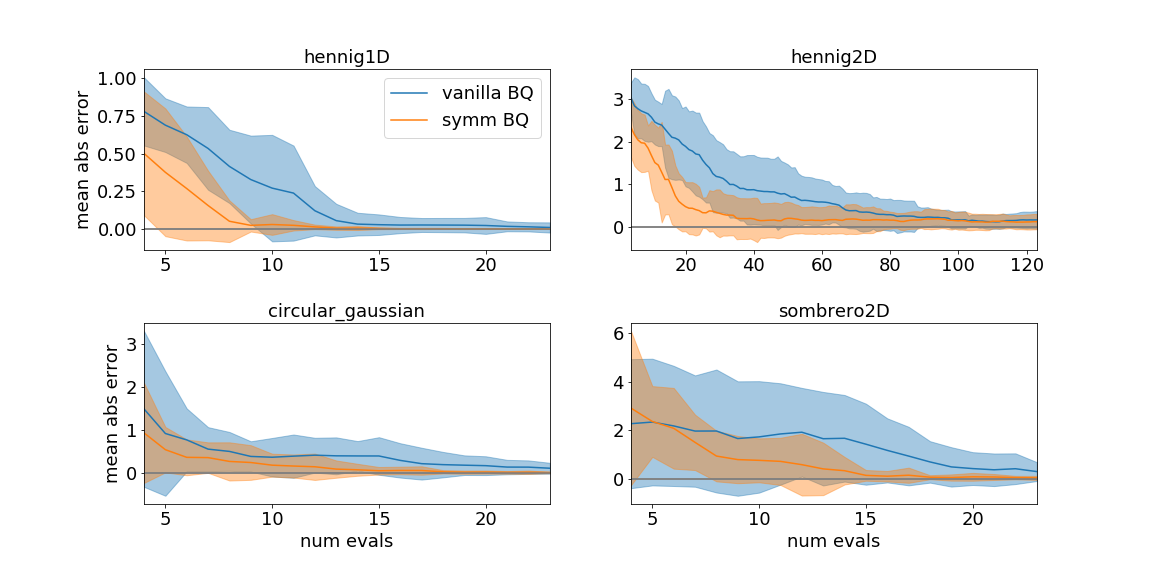}
  \caption{Plots as in  Figure~\ref{fig:symm_bq_performance_avergaes}
    but for optimal hyper-parameters $\lambda$ and $\theta$.}
  \label{fig:symm_bq_performance_avergaes_app}
\end{figure}
\begin{figure}[t]
  \centering
  \hspace*{-6mm}\includegraphics[scale=0.24]{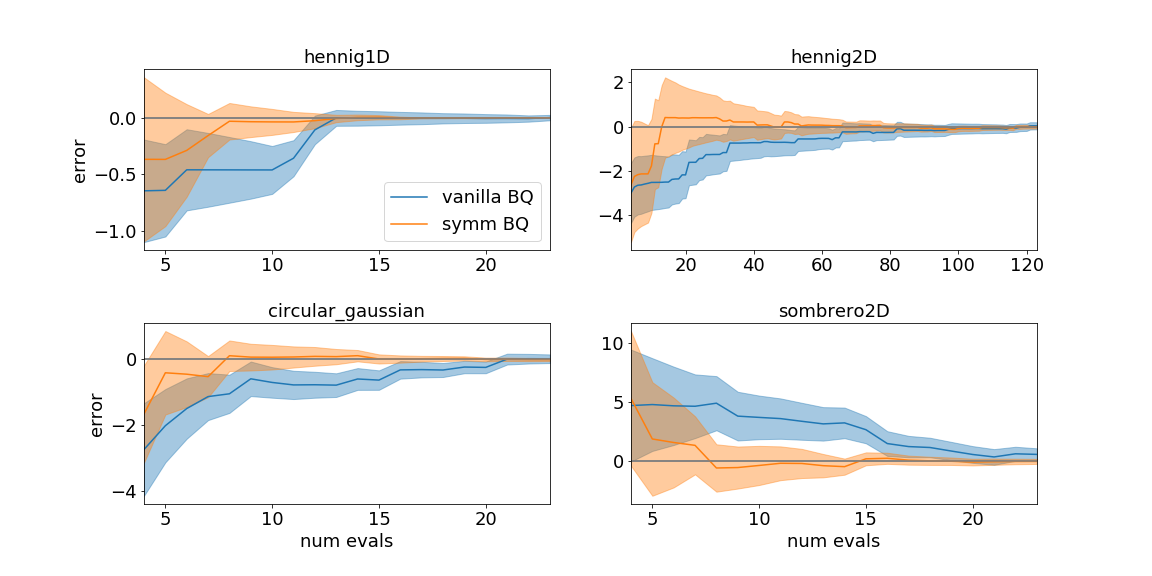}
  \caption{Plots as in  Figure~\ref{fig:symm_bq_performance_single_run}
    but for optimal hyper-parameters $\lambda$ and $\theta$.}
  \label{fig:symm_bq_performance_single_run_app}
\end{figure}

Experimental details of synthetic experiments: Observation locations are selected sequentially and actively by maximizing
the integral variance reduction acquisition function \cite{Gessner19}.
Figure~\ref{fig:symm_bq_performance_avergaes} in the main paper shows the relative mean absolute
error between the mean estimator $\mu_Z$ and the
true value of the integral, across 10 runs with different random seeds.
The runs differ in the initial design which consist of 5 randomly
chosen location in $\Omega$ and are shared by all algorithms.
Invariant-BQ uses point symmetry (orange) as well as all symmetries
(green) as in Figure~\ref{fig:Q_func}, most left and most right plot, respectively.

\begin{figure}
  \centering
  \hspace*{-6mm}\includegraphics[scale=0.30]{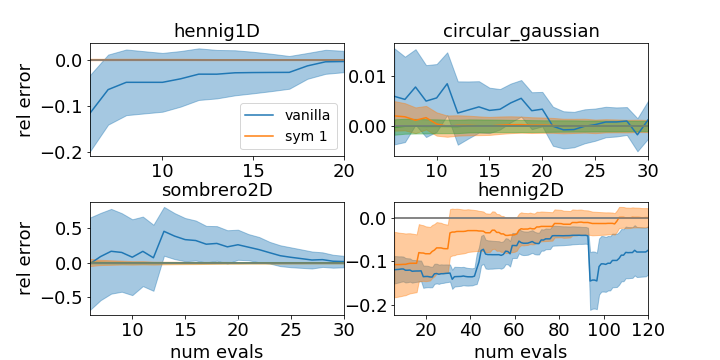}
  \caption{
    Performance for a single run taken from the runs of
    Figure~\ref{fig:symm_bq_performance_avergaes}. Details in text.
  }
  \label{fig:symm_bq_performance_single_run}
\end{figure}

Figures~\ref{fig:symm_bq_performance_avergaes_app} and
\ref{fig:symm_bq_performance_single_run_app} contain additional
results for the synthetic examples as described in
Section~\ref{sec:upperc-exampl}.
In contrast to Figures~\ref{fig:symm_bq_performance_avergaes}
and \ref{fig:symm_bq_performance_single_run} where the
kernel hyperparameters $\lambda$ and $\theta$ where set by maximizing the
marginal log-likelihood of the model, here the hyperparameters are
set to optimal values (found by over-sampling). This means that in
cases where hyper-paramters are known, the performance improvements of
invariant-BQ over standard-BQ are even more apparent.

In addition to visualising performance on synthetic examples averaged over multiple runs
in Figure~\ref{fig:symm_bq_performance_avergaes}, we plot a single run below.
Here, the solid lines show the error between $\mu_Z$ and the true integral value, and the shaded
areas show the standard deviation $\sigma_Z$ of the posterior.

\begin{figure}
  \centering
  \hspace*{-5mm}\includegraphics[scale=0.30]{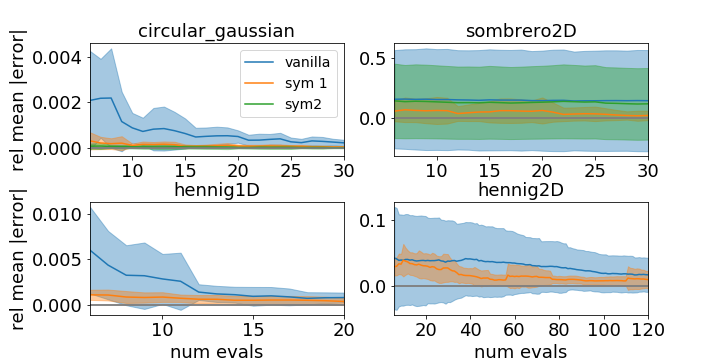}
  \caption{Same as Figure~\ref{fig:symm_bq_performance_avergaes}, but for Gaussian measure $\pi(\x)$.}
  \label{fig:symm_bq_performance_avergaes_gauss}
\end{figure}

\section*{E: Additional results of point spread function in Section~\ref{sec:upperc-spre-funct}}

\begin{wrapfigure}{r}{0.5\textwidth}
  \centering
    \includegraphics[scale=0.07]{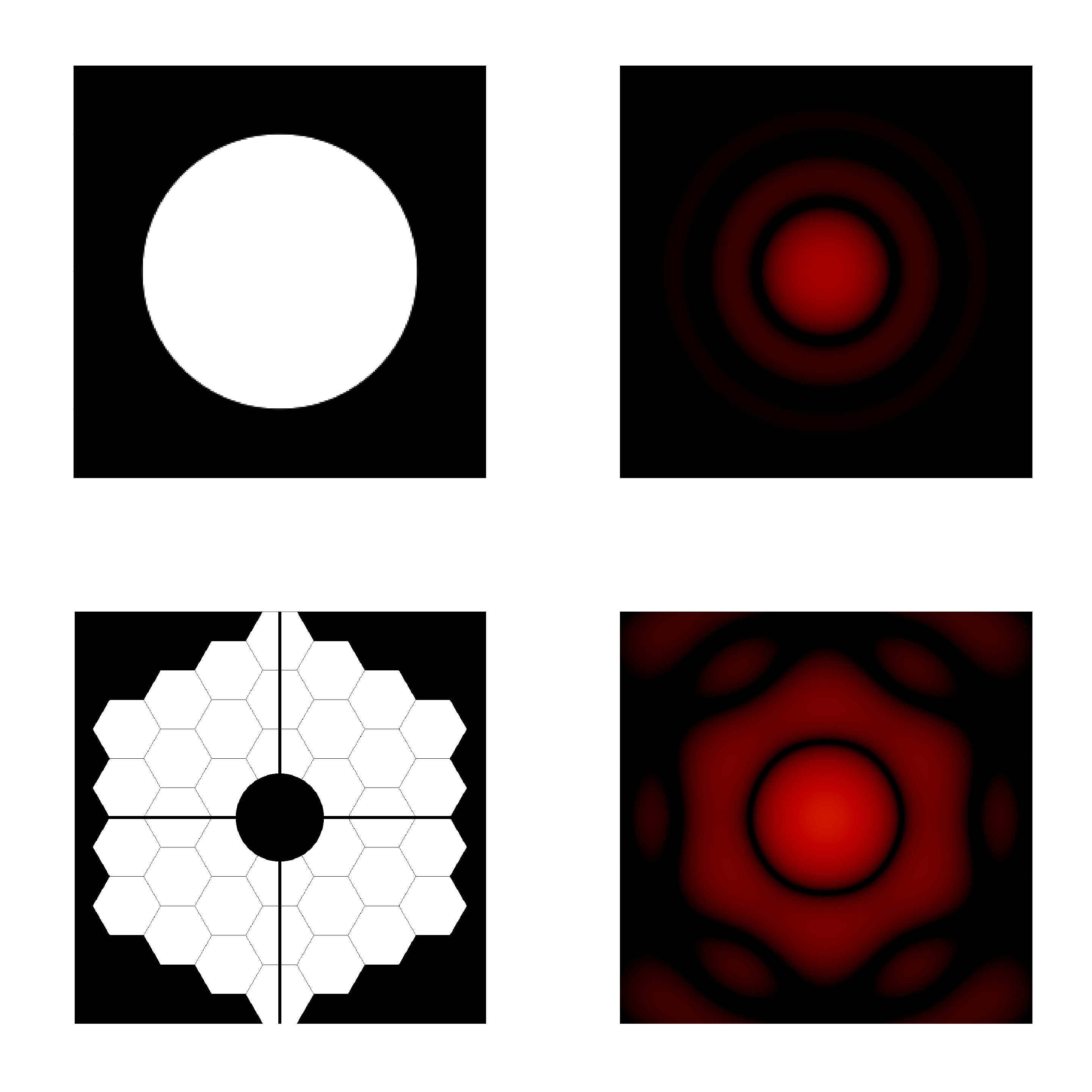}
    \caption{Pupil function (left) and PSF (right) for a perfect circular lens (top) and ATLAST (bottom).
      The PSF is computed for the wavelength $2 \times  10^{-5}$}
  \label{fig:pupil_functions_and_psfs}
\end{wrapfigure}

The point spread function (PSF) is defined as the absolute square of the Fourier
transform of the pupil function, which is a complex
function used to define the physical size and shape of the
lens.

The PSF for a perfect circular lens focused onto the point
source is known as the Airy pattern.
For a complex segmented pupil, the PSF becomes more
peculiar in its geometry, increasingly so depending on the number of
composing segments and obscurations. This can be seen in
Figure~\ref{fig:pupil_functions_and_psfs}, where we plot the pupil function and
PSF for the ATLAST pupil \parencite{postman2012advanced,
  feinberg2014cost}, a pupil composed of three rings of two-metre
segments with an added central obscuration and a spider pattern.

\end{document}